\documentclass[final]{cvpr}

\usepackage{times}
\usepackage{epsfig}
\usepackage{graphicx}
\usepackage{amsmath}
\usepackage{amssymb}
\usepackage{subfiles}
\usepackage{multirow} 
\newcommand{\ra}[1]{\renewcommand{\arraystretch}{#1}}
\usepackage{booktabs}
\usepackage[ruled]{algorithm2e}

\usepackage{makecell}

\usepackage{lipsum}

\newcommand\blfootnote[1]{%
\begingroup
\renewcommand\thefootnote{}\footnote{#1}%
\addtocounter{footnote}{-1}%
\endgroup
}

\usepackage[pagebackref=true,breaklinks=true,colorlinks,bookmarks=false]{hyperref}

\begin{document}

\title{CelebA-Spoof Challenge 2020 on Face Anti-Spoofing: Methods and Results}

\author{Yuanhan Zhang, Zhenfei Yin, Jing Shao, Ziwei Liu, \\
Shuo Yang, Yuanjun Xiong, Wei Xia, \\ 
Yan Xu, Man Luo, Jian Liu, Jianshu Li, Zhijun Chen, Mingyu Guo, \\
Hui Li, Junfu Liu, Pengfei Gao, Tianqi Hong, Hao Han, Shijie Liu, \\
Xinhua Chen, Di Qiu, Cheng Zhen, Dashuang Liang, Yufeng Jin, Zhanlong Hao}

\maketitle

\blfootnote{$\bullet$ Yuanhan Zhang is with Beijing Jiaotong University and SenseTime Research.}
\blfootnote{$\bullet$ Zhenfei Yin and Jing Shao are with SenseTime Research.}
\blfootnote{$\bullet$ Ziwei Liu is with S-Lab, Nanyang Technological University.}
\blfootnote{$\bullet$ Shuo Yang, Yuanjun Xiong, and Wei Xia are with Amazon Web
Services.}
\blfootnote{$\bullet$ Yan Xu, Man Luo, Jian Liu, Jianshu Li, Zhijun Chen and Mingyu Guo are with ZOLOZ.}
\blfootnote{$\bullet$ Hui Li, Junfu Liu, Pengfei Gao, Tianqi Hong, Hao Han and  Shijie Liu are with Meituan.}
\blfootnote{$\bullet$ Xinhua Chen, Di Qiu, Cheng Zhen, Dashuang Liang, Yufeng Jin, and Zhanlong Hao are with Vision Intelligence Center of Meituan.}

\begin{abstract}
As facial interaction systems are prevalently deployed, security and reliability of these systems become a critical issue, with substantial research efforts devoted. Among them, face anti-spoofing emerges as an important area, whose objective is to identify whether a presented face is live or spoof. Recently, a large-scale face anti-spoofing dataset, \textbf{CelebA-Spoof} which comprised of 625,537 pictures of 10,177 subjects has been released. It is the largest face anti-spoofing dataset in terms of the numbers of the data and the subjects. This paper reports methods and results in the \textbf{CelebA-Spoof Challenge 2020 on Face Anti-Spoofing} which employs the CelebA-Spoof dataset. The model evaluation is conducted online on the hidden test set. A total of 134 participants registered for the competition, and 19 teams made valid submissions. We will analyze the top ranked solutions and present some discussion on future work directions.

\end{abstract}

\section{Introduction}
Face interaction systems have become an essential part in real-life applications, with the successful deployments in electronic identity authentication. 
Meanwhile, it is challenging to deal with Presentation Attacks (PA)~\cite{PA1} in practical usage. 
In order to protect our privacy and property from being illegally used by others, 
Face Anti-Spoofing (FAS)~\cite{PA2,FAS1,AtoumFaceAU}, which aims to determine whether a presented face is an attacker or client, has emerged as a crucial technique and attracted extensive interests in recent years~\cite{Galbally2014FASsurvey}.

Leveraging on the CelebA-Spoof~\cite{CelebA-Spoof} dataset, we organize the \textit{CelebA-Spoof Challenge 2020 on Face Anti-Spoofing} (CelebA-Spoof Challenge) collocated with the Workshop on Sensing, Understanding and Synthesizing Humans at ECCV2020~\footnote{Workshop website: \url{https://sense-human.github.io/}.}. The goal of this challenge is to boost the research on face anti-spoofing. Specifically, the CelebA-Spoof is comprised of 625,537 pictures of 10,177 subjects, which is the largest face anti-spoofing dataset in terms of the numbers of the data and the subjects. The dataset also features a hidden test set containing around 30000 images for online evaluation of this challenge. The dataset construction of the hidden dataset is the same as the public dataset.

In the following sections, we will describe this challenge, analyze the top ranked solutions and provide discussions to draw conclusion derived from the competition and outline future work directions.

\section{Challenge Overview}

\subsection{Platform}
The CelebA-Spoof Challenge is hosted on the CodaLab platform~\footnote{Challenge website: \url{https://competitions.codalab.org/competitions/26210}.}. After registering on the CelebA-Spoof Challenge, each team is allowed to submit their models to the Amazon Web Services (AWS)~\footnote{Online evaluation website: \url{https://aws.amazon.com}.}, and each team is allocated one 16 GB Tesla V100 GPU to perform online evaluation on the hidden test set. The encrypted prediction files including results of each data in hidden test set are sent to the teams through automatic email when their requested online evaluation has done. Teams are required to upload their encrypted prediction files to the CodaLab platform for the ranking. 

\subsection{Dataset}
The CelebA-Spoof Challenge 2020 on Face Anti-Spoofing employs the CelebA-Spoof dataset~\cite{CelebA-Spoof} that was proposed in ECCV 2020. CelebA-Spoof is a large-scale face anti-spoofing dataset that has 625,537 images from 10,177 subjects, which includes 43 rich attributes on face, illumination,environment and spoof types. Live image selected from the CelebA dataset~\cite{CelebA}. We collect and annotate spoof images of CelebA-Spoof. Among 43 rich attributes, 40 attributes belong to Live images including all facial components and accessories such as skin, nose, eyes, eyebrows, lip, hair, hat, eyeglass. 3 attributes belong to spoof images including spoof types, environments and illumination conditions.CelebA-Spoof can be used to train and evaluate algorithms of face anti-spoofing. The hidden test set is devised for the CelebA-Spoof Challenge, the data construction of the hidden test is as the same as the public test set. All the teams participating the CelebA-Spoof Challenge are restricted to train their algorithms on the publicly available CelebA-Spoof training dataset.

\subsection{Evaluation Metric}
Considering face anti-spoof as binary classification, we can leverage FPR@TPR as evaluation criteria. Specifically, spoof class is Positive, live class is Negative.
\begin{equation}
    FPR = \frac{FP}{FP+TN}, \quad TPR = \frac{TP}{TP+TN} 
\end{equation}

Specifically, \textit{FPR}, \textit{TPR}, \textit{TP}, \textit{TN} and \textit{FP} correspond to False Positive Rate, True Positive Rate, True Positive, True Negative and False Negative. The TPR@FPR=$\text{5}^\text{-3}$ determines the final ranking. Besides, we also provide TPR@FPR=$\text{10}^\text{-3}$ and TPR@FPR=$\text{10}^\text{-4}$. Among all the uploaded results, if their TPR@FPR=$\text{5}^\text{-3}$ is the same, the one with higher TPR@FPR=10E-4 will achieve a higher ranking.

\subsection{Timeline}
The CelebA-Spoof Challenge lasted for nine weeks from August 28, 2020 to October 31, 2020. During the challenge, participants had access to the public CelebA-Spoof dataset, and they are restricted to used the public CelebA-Spoof training dataset for the training of their model. The challenge results were announced in February 10, 2021. A total of 134 participants registered for the competition, and 19 teams made valid submissions.

\section{Results and Solutions}
Among the 19 teams who made valid submissions, many participants achieve promising results. We show the final results of the top-5 teams in Table~\ref{table:all_result}. In the following sections, we will present the solutions of the top-3 entries.

\setlength{\tabcolsep}{5pt}
\begin{table}[t]
\centering
\ra{1.1}
\caption{Final results of the top-5 teams in the CelebA-Spoof Challenge 2020 on Face Anti-Spoofing.}
\vspace{3pt}
\label{table:all_result}
\resizebox{0.48\textwidth}{!}{%
\begin{tabular}{ccccccccc}
\Xhline{1pt}
\multirow{2}{*}{Ranking} & \multirow{2}{*}{TeamName} & \multirow{2}{*}{UserName} & \multicolumn{3}{c}{TPR (\%)$\uparrow$}  \\ \cmidrule{4-6}
 &  & & FPR=$\text{10}^\text{-3}$ & FPR=$\text{5}*\text{10}^\text{-3}$ & FPR=$\text{10}^\text{-6}$ \\  \hline
1       & ZOLOZ    & ZOLOZ    & 1.00000       & 1.00000      & 1.00000         \\
2       & MM       & liujeff  & 1.00000       & 1.00000      & 0.99991        \\
3       & AFO      & winboyer & 1.00000       & 1.00000      & 0.99918         \\
4       & RealFace & zkyezhang      & 0.99991      & 0.99936      & 0.92268         \\
5       & k\_   	   & k\_       & 0.99973      & 0.99927      & 0.98026        \\ \Xhline{1pt}
\end{tabular}%
}
\end{table}

\subsection{Solution of First Place}
\textit{Team members: Yan Xu, Man Luo, Jian Liu, Jianshu Li, Zhijun Chen, Mingyu Guo}
\begin{figure}[h!]
\centering
\includegraphics[width=0.45\textwidth]{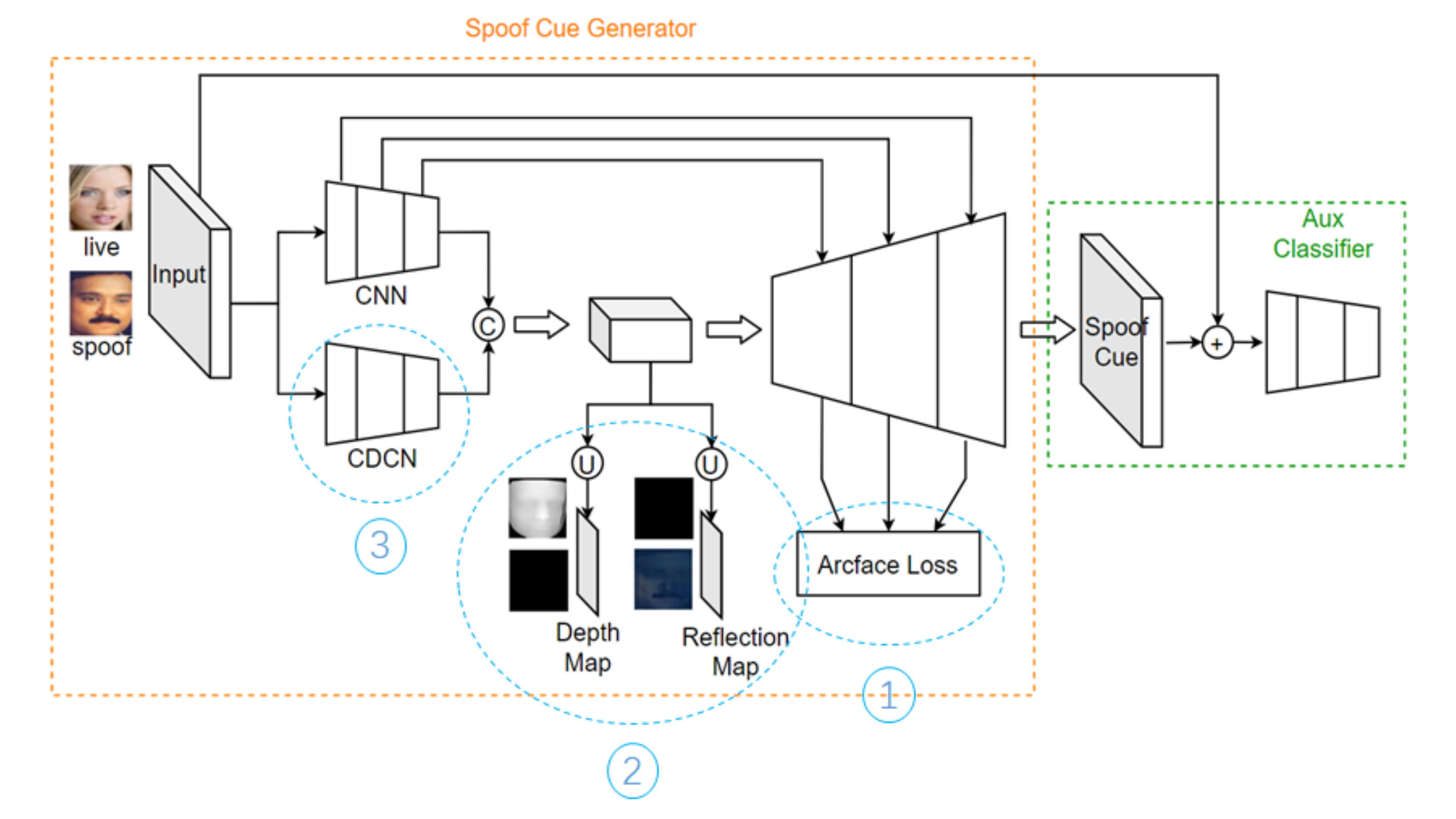}
\caption{The framework of the first-place solution.}
\label{figure:the first place}
\end{figure}

\noindent \textbf{General Method Description.}
The champion team propose a robust method for face anti-spoofing. It
has two components:
\textbf{1) Spoof modeling}: in which they adopt several state-of-art models to predict spoof cue of each testing image.
\textbf{2) Spoof fusion}: in which they propose a heuristic voting strategy for robust multiple scores combination.

\noindent \textbf{Spoof Modeling.}
In order to obtain the spoof cue for the attacked images, they combine a bag of start-of-the-art models to find the spoof evidence of each testing image in this competition. Specifically, they propose a novel framework named FOCUS (\textbf{F}inding sp\textbf{O}of \textbf{CU}e for face anti-\textbf{S}poofing) to handle with face anti-spoofing problem. A multi-task learning based model AENet~\cite{CelebA-Spoof}, a binary task based model ResNet~\cite{ResNet} and a attack types classification base model are adopted to enhance the ability to detect the spoof cue. Furthermore, a noise print method is adopted to recognize the device type of attacked image.

\noindent $\bullet$ \textit{FOCUS:} 
As shown in Figure~\ref{figure:the first place}, inspired by~\cite{CelebA-Spoof,feng2020learning}, they propose a novel framework FOCUS for face anti-spoofing. It mainly includes two modules: a Spoof Cue Generator and an Aux Classifier. The spoof cue generator adopts the U-Net structure with an encoder and a decoder to generate spoof cue of the same size as the input image. Regression loss is utilized during training process to minimize the spoof cue of live images. Meanwhile, it does not apply any constraints on the spoof images. In order to improve the generalization ability of unknown attack types, they design a two-path encoder and adopt ResNet18\_CDC~\cite{yztCDC} as the backbone of each encoder. In addition, they introduce a reflection map~\cite{zhang2018single} and a depth map~\cite{KimBASN} in the latent space of the encoder, and adopt 3D geometric information as an auxiliary constraint. As a result, the features of the latent space will have higher responses on the spoof image. In the decoder part, they introduce multi-branch arcface loss~\cite{deng2018arcface} to improve the compactness within the live class and the distinction bettheyen the live-spoof classes. For the auxiliary classifier, they design a binary classification model connected after the generator to assist the end-end training of the whole framework.

\noindent $\bullet$  \textit{AENet:} 
AENet~\cite{CelebA-Spoof} is adopted to predict the spoof score of each testing image.

\noindent $\bullet$ \textit{ResNet:}
Through bad-case analysis, they find that AENet is not good at detecting spoof images for mask and outdoor scenarios. To this end, they deploy a binary classification model ResNet-18~\cite{ResNet} to enhance the spoof-detecting  ability. During the training status, the training samples of spoof images are only from masks and outdoor attacks. Furthermore, focal loss is adopted to solve the over fitting problem from easy samples. Meanwhile, a series of data argumentation strategies, such as random crop, image flip and color distortion, are adopted to improve its generalization ability.

\noindent $\bullet$ \textit{Attack types:}
By analyzing the CelebA-Spoof train data, they find that the different spoof images have similar attack clues, such as similar display borders, similar backgrounds, and similar paper printing edges. To this end, they train a attack-type based model to predict the attack clue. Specifically, they first remove the foreground area which containing face context, since the spoof clue is the feature in a attack image. Then they train a classification module on various spoof types.

\noindent $\bullet$ \textit{Noise Print:}
Different camera's digital imaging pipeline have common processes like  data compression, interpolation, and gamma correction, and also have unique processes to offer more advanced functionalities. The unique process varies from camera model to camera model, and the acquired images from different cameras have artifacts which are peculiar to  the camera itself, and hence can be used to perform face anti-spoof task. In this competition, One feature has been observed from train set, the live images are collected from internet or social medial while the spoof images are directly captured from device cameras, e.g. Phone camera, Pad camera or PC camera. they find that the different noise prints on different device cameras, therefore they utilize noise print as a feature to represent the camera type. To extract the noisy print, they fist apply DCT transformation and quantization on this image, then total 64 frequency density histograms are calculated based on 8x8 macro block of DCT coefficients. For each frequency density histogram, FFT is applied to obtain the the number of peaks that is exceeding the pre-defined threshold t. Finally, they can use 64 dimensional vector to represent noise print of different camera type. During training procedure, they first divide the train set as four groups, group 1 from live images, group 2 from spoof images of Phone, group 3 from spoof images of Pad and group 4 from spoof images of PC. Then noise prints of four group are extracted and send them to a network to distinguish the different distribution of each noise print.

\noindent \textbf{Spoof Fusion.}
To obtain the best performance of TAR (True accepted rate) at given FAR (False accepted rate) in the face anti-spoofing task, they propose a heuristic voting scheme at the score level for robust combinations of different models. they first normalize all confidence scores of each trained
model to 0-1, and they assign the model with the best performance as the main model while they regard the others as auxiliary models. Then, they change the confidence score to 0 or 1 if all models have similar prediction ranges. The scores are amended to 0 or 1 if other auxiliary models have strong confidence of belonging to live face or spoofing face, respectively. For images whose scores are not close to 0 or 1, they consider them as hard cases, because they are laying on the edge of decision boundaries of each model. For these cases, they re-arrange the scores to around 0.1.

\noindent \textbf{Implementation details:}
For fusion strategy, they adopt a heuristic voting scheme to obtain the best performance on TAR and FAR, please refer to “Testing description” for more details. The fusion strategy allow us to achieve $100\%$ TAR when FAR is $\text{5}*\text{10}^\text{-3}$ and $\text{10}^\text{-4}$. 

FOCUS is implemented with Pytorch and trained end-to-end. In the training stage, models are trained with Adam optimizer and the initial learning rate (lr) and the weight decay (wd) are $\text{2}*\text{10}^\text{-4}$ and $\text{5}*\text{10}^\text{-5}$, respectively. they train models with maximum 25 epochs while lr decays every 6 epochs by a factor of 0.3. During training, the training samples are resampled to keep the live-spoof ratio as close to 1:1. The training batch size is 64 on four 1080Ti GPUs. they initialize the backbone ResNet18\_CDC in the encoders with MSRA method. Besides, their pipeline spend 24h for training/0.8s for testing each image (not including the pre-processing) 

\subsection{Solution of Second Place}
\textit{Team members: Hui Li, Junfu Liu, Pengfei Gao, Tianqi Hong, Hao Han, Shijie Liu}
\begin{figure}[h!]
\centering
\includegraphics[width=0.45\textwidth]{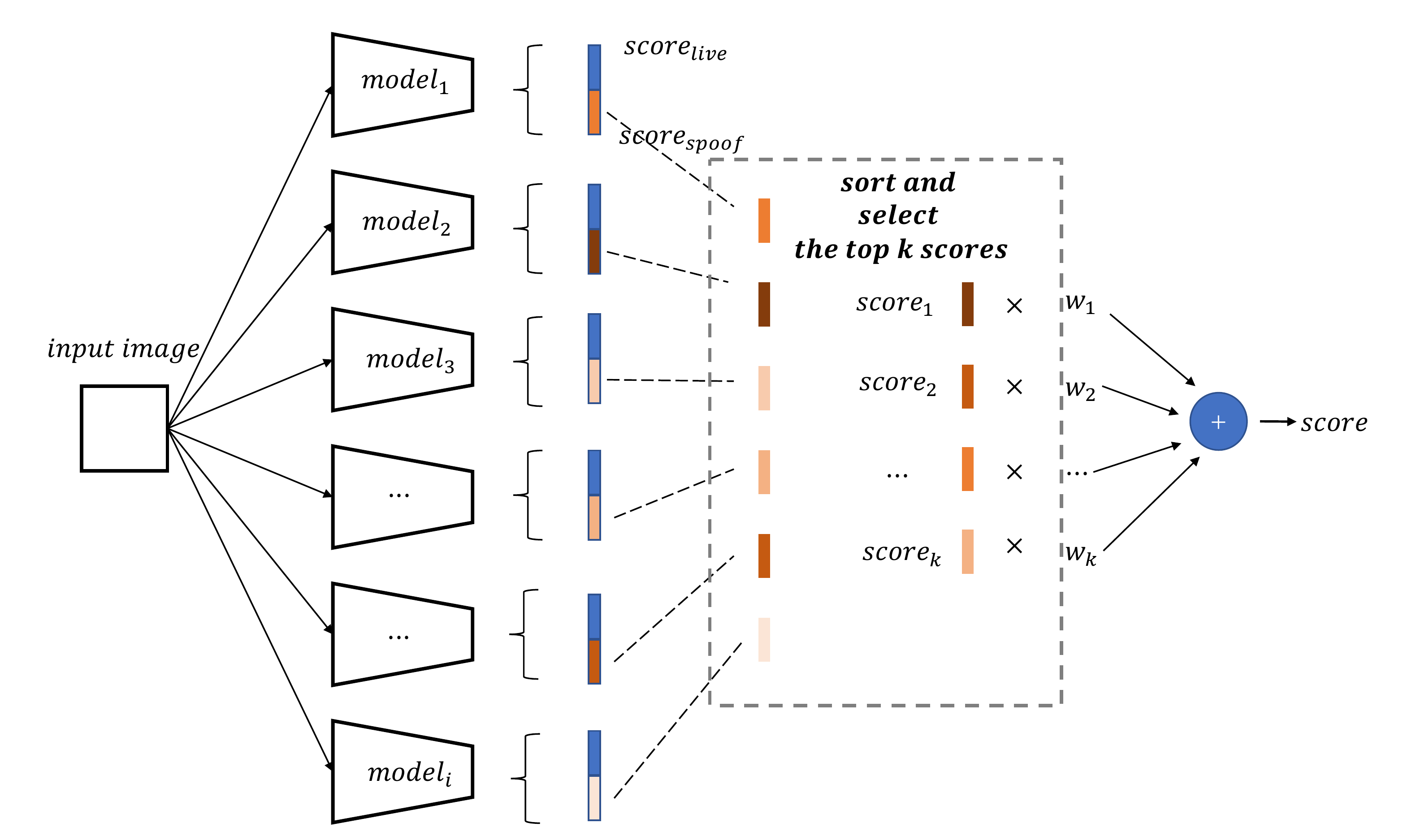}
\caption{The framework of the second-place solution.}
\label{figure:the second place 1}
\end{figure}

\noindent \textbf{General Method Description.}
In this challenge, they adopt five different models and ensemble with a ``weight-after-sorting" strategy for face anti-spoofing. In the training and test stage of several networks, they use patches to keep the model focusing on the spoof cues instead of other irrelevant face features, which makes the trained networks more robust and generalizable. In the fusion stage, they propose a novel ensemble strategy, naming ``weighting-after-sorting". The output scores of different methods will first be sorted and they select the top $k$ scores and assign which with different weights, searched by the Particle Swarm Optimization (PSO) algorithm. This strategy is rank-specific instead of model-specific, which further enhances the performances of their method.

\noindent \textbf{Training Description.}
They used 5 single models for the further model ensemble. The details of the single models are as follows. 

\noindent $\bullet$ \textit{CDCNpp:} They used the Central Difference Convolutional Network~\cite{yztCDC}. Instead of training on the whole image, they used random patches of the face images as inputs. They trained the CDCNpp on two scales of patches, 64*64 and 96*96 respectively. They adopted grayscale images as the depth supervision in CDCNpp, the sizes of the grayscale images are 16*16 and 24*24 respectively.

\noindent $\bullet$ \textit{LGSC:} They adopted the LGSC~\cite{feng2020learning}. The input images were resized to 224*224 and the model was initialized with the pre-trained ResNet18 model. They used batch balance sampler to balance the positive samples and negative ones in a batch.

\noindent $\bullet$ \textit{SeResNet50:} They adopted SeResNet50 for a simple binary classification. They used images resized to 224*224 as inputs and the pre-trained SeResNet50 model.

\noindent $\bullet$ \textit{EfficientNet-b7:} All settings~\cite{tan2019efficientnet} are same as the SeResNet50, with inputs with sizes of 224*224 and a pre-trained model on ImageNet.

\noindent $\bullet$ \textit{SeResNeXt50:} The training took random patches with sizes of 64*64 as inputs. To take advantage of other supervision information like the spoof types and the illumination types provided in the training set, they adopted a multi-task learning similar to the AENet~\cite{CelebA-Spoof} and added two fully connected layers in the tail of SeResNeXt50, predicting the spoof types and the illumination types respectively. The losses are all softmax cross entropy losses, the weights of spoof types loss and the illumination types loss are set to 0.1 and 0.01 respectively.

Besides, the fusion strategy and the image transforms used by the above-mentioned methods are shown in Fig~\ref{figure:the second place 1} and Table~\ref{table:the second place 1}.

\begin{figure}[t]
\centering
\includegraphics[width=0.45\textwidth]{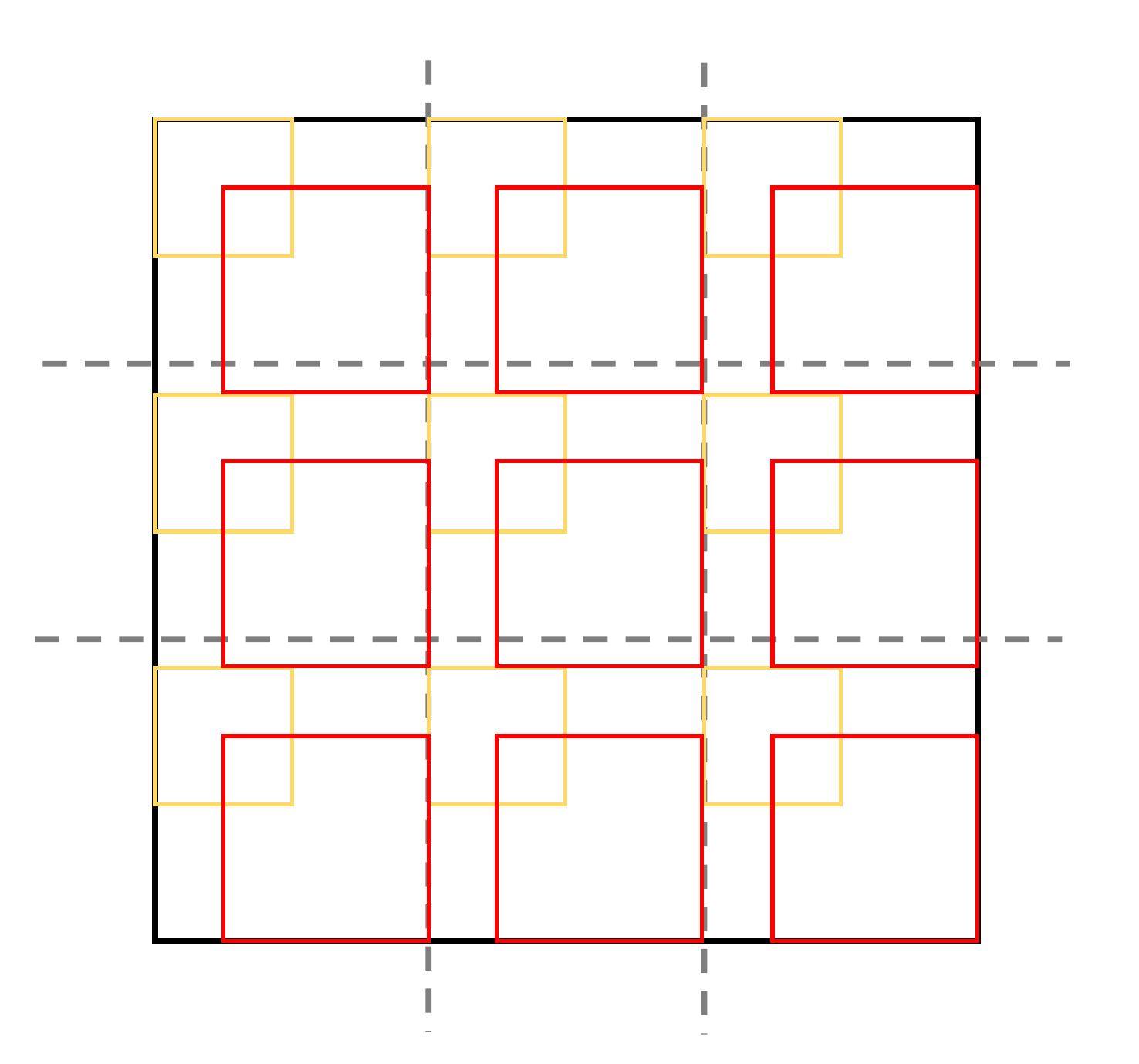}
\caption{The framework of the second-place solution.}
\label{figure:the second place 2}
\end{figure}

\setlength{\tabcolsep}{5pt}
\begin{table}[t]
\centering
\ra{1.1}
\caption{Image transforms in the training stage of the second place solution.}
\vspace{3pt}
\label{table:the second place 1}
\resizebox{0.48\textwidth}{!}{
\begin{tabular}{c|ccccc}
\Xhline{1pt}
Methods  & CDCNpp & LGSC & SeResNet50 & EfficientNet-b7 & SeResNeXt50 \\\hline
RandomHonrizontalFlip & \checkmark & \checkmark & \checkmark & \checkmark &\checkmark\\
RandomRotation & \checkmark & \checkmark & \checkmark & \checkmark &\checkmark \\
RandomErasing & \checkmark &  &  &  & \\
Cutout &\checkmark &  & \checkmark & \checkmark &\checkmark \\
ColorJitter & \checkmark & \checkmark & \checkmark & \checkmark &\checkmark \\
Mixup &  &  & \checkmark & \checkmark & \\ \Xhline{1pt}
\end{tabular}
}
\end{table}

\noindent \textbf{Testing Description.}
The testing strategies of the above single models are as follows:

\noindent $\bullet$ \textit{CDCNpp}: The input image was first split into 3*3 parts and they crop the upper left corner with size of 64*64 and the lower right corner with size of 96*96 of each part to generate test patches of the two CDCNpp trained on different size of patches. Figure~\ref{figure:the second place 2} illustrates how they generate the patches. They applied the horizontal flip to the 96*96 patches. For each CDCNpp, they calculated the mean value of the nine patches of a test image as the prediction score of the image.

\noindent $\bullet$ \textit{SeResNeXt50}: They cropped the center part with size of 64*64 of the image and flip the patch horizontally. The prediction score of SeResNeXt50 is calculated as the mean value of the two patches. 

\noindent $\bullet$ \textit{Others}: For other models mentioned above, the inputs images are resized to 224*224, then sent to the LGSC, the SeResNet50 and the EfficientNet-b7.

\noindent \textbf{Implementation details.}
As has been introduced above, they tested an image with six models (two CDCNpp with different sizes of patches and four other models) and get six scores of the input image belonging to the spoof class. They proposed a novel ``weighting-after-sorting" strategy for the model ensembles. Specifically, they first sorted six scores in descending order, then selected the top $k$ scores for score fusion. They used the Particle Swarm Optimization (PSO) algorithm to find the $k$ weights assigned to the top $k$ scores at different ranks with the best performance on the validation set. It can be seen with that such strategy, the weights are not model-specific, but rank-specific. Considering a spoof image with only one model giving high score, such model may not weight enough in previous model-specific fusion strategy, but its score will certainly be noticed in their weighting-after-sorting strategy. The $k$ was set to 4 in their final submission. Besides, their pipeline spend 18h for training/0.076s for testing each image (pre-processing included).

\subsection{Solution of Third Place}
 \textit{Team members: Xinhua Chen, Di Qiu, Cheng Zhen, Dashuang Liang, Yufeng Jin, Zhanlong Hao}
 
\begin{figure}[h!]
\centering
\includegraphics[width=0.45\textwidth]{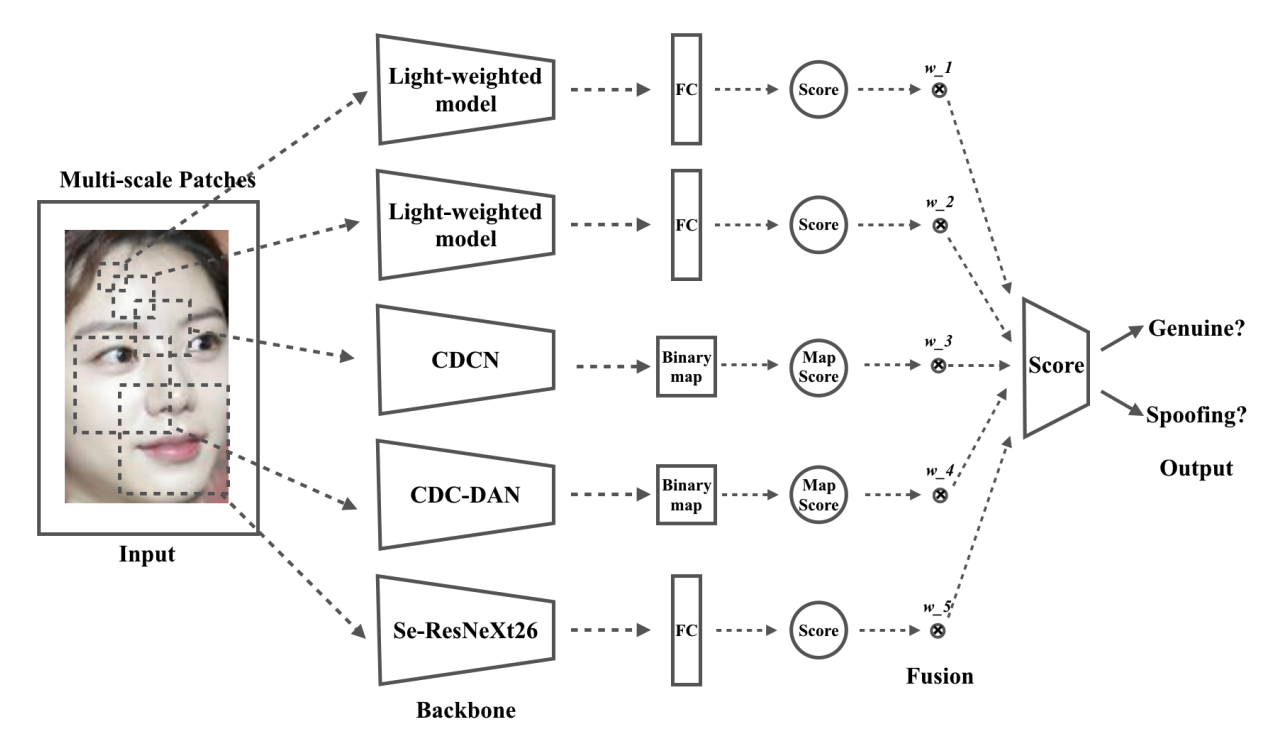}
\caption{The framework of the third-place solution.}
\label{figure:the third place 1}
\end{figure}

\begin{figure}[t]
\centering
\includegraphics[width=0.45\textwidth]{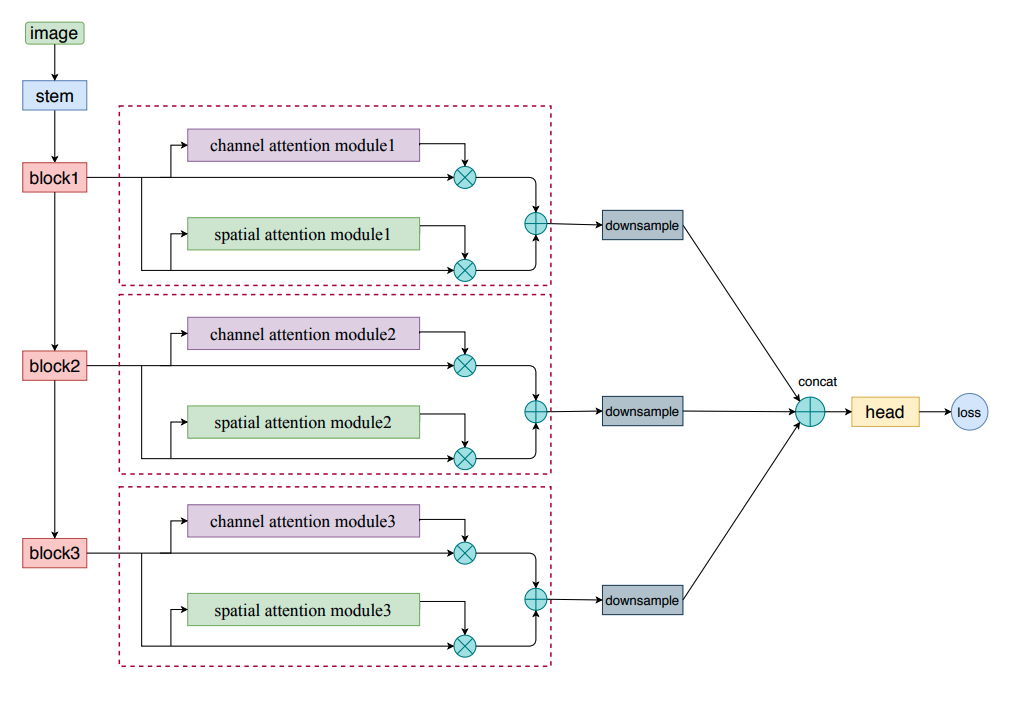}
\caption{The backbone of the purposed CDC-DAN.}
\label{figure:the third place 2}
\end{figure}

\noindent \textbf{General Method Description.}
AS shown in Fig~\ref{figure:the third place 2}, they propose a novel method based on fusing CDCN~\cite{yztCDC} and DAN~\cite{fu2019dual}. CDCN is able to capture detailed patterns via aggregating both intensity and gradient information, while DAN with self-attention mechanism can enhance the discriminated ability of feature representations through spatial inter-dependencies and channel inter-dependencies. When combined, they can significantly improve the face anti-spoofing performance by modeling rich contextual information over local intensity and gradient features. In addition, motivated by the insight that artificial spoofing information of the image is independent on semantic information, they utilize facial patches~\cite{AtoumFaceAU} as the input for models that they tend to decouple the spoofing feature from its full-face feature. The full-face image is divided into many different facial patches which can force the proposed networks to focus on the spoof-specific discriminative information. Finally, the introduced multi-scale strategy aims at generating multi-scale patches from the original image. Random crop is applied to produce patches on the different scales that they use five scales of patches, i.e., 32*32, 48*48, 64*64, 112*112, and 128*128 respectively as the input size for the proposed CNN-based networks. In general, they propose a multi-scale patch-based CDC-DAN method for face anti-spoofing detection.

\noindent \textbf{Training Description.}
In the training stage, to reduce overfitting, convolutional neural networks are typically trained with data augmentation. For face anti-spoofing, they find that mixed-example data augmentation is very useful. they utilize a variety of methods including cutout, vh-mixup, mixed-concat, random square and random interval, to generate mixed-example data augmentation, which result in improvements over models trained without any form of mixed-example data augmentation. Moreover, if the image size is smaller than the preset input size, they enlarge the image by mirroring instead of scaling, which greatly improves the performance.

\noindent \textbf{Testing Description.}
In the test stage, they do not use any type of data augmentation methods. They only adjust the image size by mirroring to meet different network input requirements. Then, they uniformly sample multiple small patches of different sizes and input them into the corresponding neural network models. Finally, they ensemble the output results of different patches and different models.

\noindent \textbf{Implementation details.}
As shown in Fig~\ref{figure:the third place 2}, their fusion strategy is a simple method that adjusts the weights of CNN
models based on the best performance of the validation dataset. The main benefit of using fusion is that the performance of average prediction is better than any contributing member in the fusion. The mechanism for improving performance through fusion is usually the reduction of the variance of the predictions made by each individual model. Besides, CDCN-DAN spends 2 days for training with 8 GPUs, Se-resnext26 spends 12 hours for training with 8 GPUs, and Light-weighted Network spends 4 hours for training with 4 GPUs.

\section{Discussion}
The winning solutions mentioned above have achieved promising results on the CelebA-Spoof Challenge, these solutions focus on different aspects in developing a robust and efficient face anti-spoofing model. To briefly sum up, among their solutions, there are two key points are essential for improving the performance of the face anti-spoofing task. \textit{1) Spoofing Cues Model:} Besides the commonly used deep learning models, such as ResNet and EfficientNet, these solutions not only inherit the models which are published recently~\cite{feng2020learning,yztCDC}, but also devise novel framework for detecting spoofing cues, such as attack-type based model, Noise Print based model as mentioned in the first place solution, and CDC-DAN proposed by the third place solution. \textit{2) Ensemble Strategy:} These winning methods leverage on different ensemble strategies to boost their model performance, such as the heuristic voting scheme of the first place solution and “weight-after-sorting” strategy of the second place solution. Moreover, we believe that there is still much room for improvement in the future face anti-spoofing challenge.  For example,  \textit{1) Size:} The size of hidden set could be larger in the future.  \textit{2) Diversity:} The live images could be more realistic instead of inheriting from the CelebA~\cite{CelebA}.
\vspace{8pt}

\noindent \textbf{Acknowledgments.} We thank Amazon Web Services for
sponsoring the prize of this challenge. Besides, we sincerely thank the codebase from DeeperForensics Challenge~\footnote{\url{https://competitions.codalab.org/competitions/25228}.}, especially for the helpful discussions from Zhengkui Guo and Liming Jiang.


{\small
\bibliographystyle{ieee_fullname}
\bibliography{challenge/sections/bibliography}
}

\end{document}